# Designing AI Pipelines for Decision-Ready ITSM Intelligence

*Short Paper*

**Archan Dutta**
Automation Anywhere

**Yash Dharmadhikari**
Automation Anywhere

**Marat Valiullin**
Automation Anywhere

**Rahul Guha**
Automation Anywhere

**Dr. Alexander Liss**
Automation Anywhere

## Introduction

IT service management (ITSM) systems record large volumes of incidents, requests, comments, and resolutions. These data are valuable not only for operational service delivery but also as traces of recurring organizational pain points, process bottlenecks, and capability gaps. Raw ITSM exports are difficult to convert into decision-ready intelligence because they are schema-inconsistent, text-heavy, noisy, and highly granular. This challenge is especially visible when non-technical stakeholders attempt to use ticket data. Sales Engineers and executives typically do not need thousands of raw tickets. Instead, they need a compact, interpretable view of recurring business-relevant issues, paired with enough drill-down detail to substantiate a customer conversation. This paper identifies three research gaps and answers three research questions (RQ) corresponding to each gap:

1. Process-design gap: Organizations lack schema-agnostic processes to convert heterogeneous ITSM exports into decision-ready intelligence without extensive manual effort
   - RQ1: What process design enables schema-agnostic transformation of heterogeneous ITSM exports into a multilevel decision-support report?
2. Abstraction gap: Limited guidance exists on how granular ticket clusters can be transformed into usable sales/executive themes.
   - RQ2: How can granular ticket clusters be abstracted into coherent and distinct themes?
3. Decision-support gap: Even when AI systems produce analytic outputs, whether business stakeholders find those outputs managerially meaningful and adoption-worthy remains under examined.
   - RQ3: To what extent do stakeholders perceive the generated outputs as useful, actionable, trustworthy, and likely to be used?

This paper makes three contributions:

1. It presents an implemented AI pipeline combining GenAI and classical ML for transforming ticket data into multilevel intelligence.
2. It outlines a hybrid ML-human evaluation strategy for abstraction quality.
3. It repositions ITSM analytics from a purely operational exercise to a sociotechnical IS problem of data transformation, abstraction, and managerial use, by establishing stakeholder-centered metrics.

## Literature Review

Design science research (DSR) frames artifact creation and evaluation as a core mode of IS knowledge production, with growing integration of machine learning into decision-support artifacts and dashboards (Hevner et al., 2004; Onwujekwe & Weistroffer, 2025). Within ITSM, prior work focuses on operational analytics: clustering and labeling tickets (Roy et al., 2016), fine-grained incident labeling (Liu et al., 2023), generative assistants for incident handling (Schmidt et al., 2024), and text mining for services management (Kumar et al., 2021). To address the process gap (RQ1), we draw on current ticket analytics processes in the literature. HCLTech reports that auditing ~12,000 ServiceNow tickets per month required ~15 minutes per ticket, consuming roughly 900 person-hours monthly. Teams spend 70% of analytic effort on data preparation, consistent with estimates that data scientists spend up to 80% of their time on extraction and cleaning (Fernandes et al., 2023).

To address the abstraction gap (RQ2), we draw on clustering validity and topic modeling literature. While HDBSCAN supports nested structures suited to noisy corpora (Campello et al., 2015), classical validity measures such as silhouette width and Davies-Bouldin index capture only geometric cohesion (Rousseeuw, 1987; Davies & Bouldin, 1979). Topic modeling research shows that statistical quality alone is insufficient, and human usefulness depends on semantic coherence and interpretability (Newman et al., 2010). Explainable clustering further argues that decision-facing results must be transparent and justifiable to human users (Álvarez-García et al., 2024). RQ2 therefore adopts a hybrid logic combining ML-based metrics with stakeholder judgments on cluster quality. The decision-support gap in RQ3 draws on IS success, technology acceptance, task-technology fit, dashboard design, and AI trust. Task technology fit (Goodhue & Thompson, 1995) and dashboard research (Yigitbasioglu & Velcu, 2012) support the value of concise, structured displays for managerial decision-making. Trust is treated as distinct from usefulness, as it materially shapes adoption and reliance on AI artifacts (Glikson & Woolley, 2020). RQ3 therefore evaluates interpretability, actionability, trust, and likelihood of use as stakeholder centered indicators of decision-support value.

Existing work remains largely operational in orientation, focused on classification, routing, or incident handling. This paper addresses the managerial gap by treating ITSM-to-report transformation as a design science problem rather than solely a text-mining problem.

## AI-Enabled Process Design

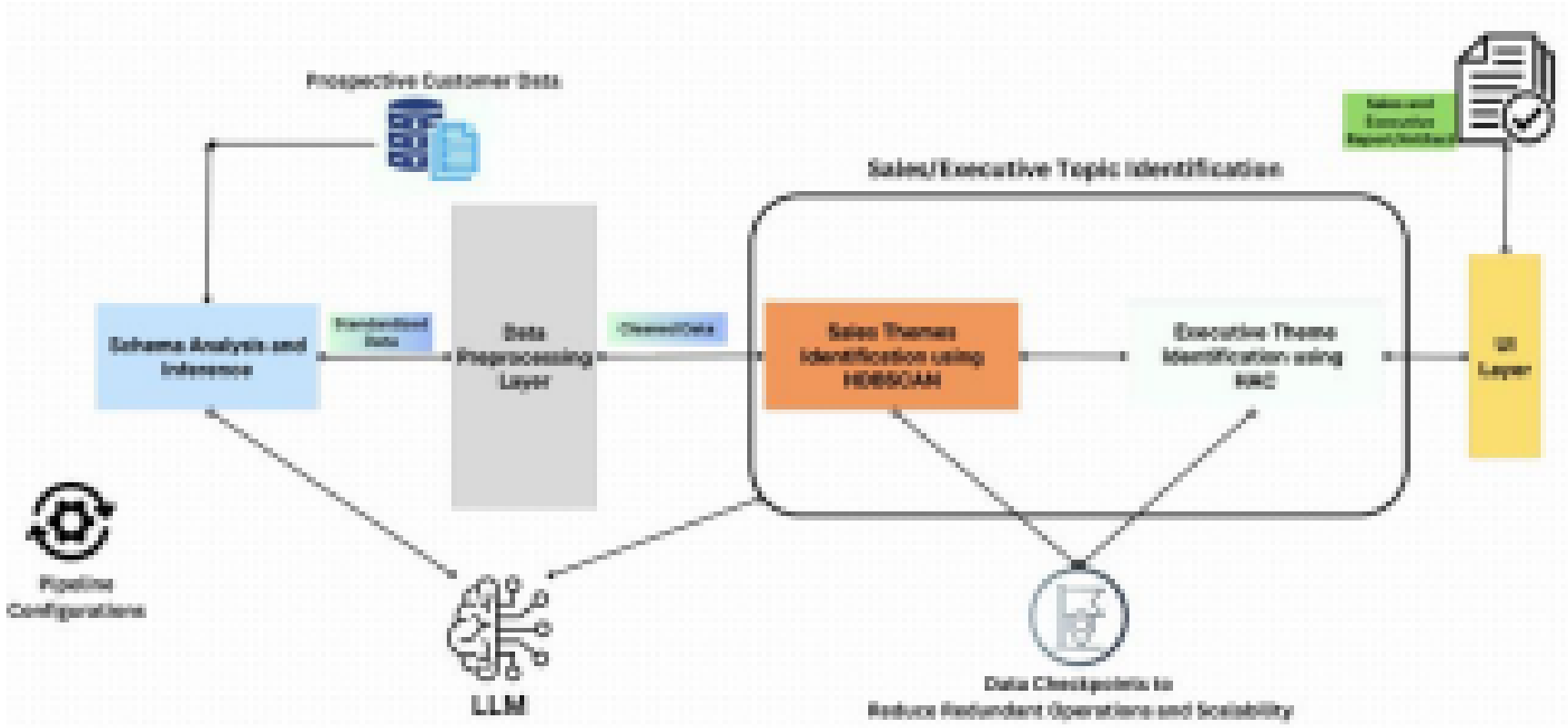

Figure 1 - High Level Architecture of the Pipeline

Following DSR logic, the process addresses RQ1 and is built around four design objectives: standardize heterogeneous ticket exports, discover recurrent themes, abstract them into managerial themes, and present them in a report suitable for sales and executive use. This AI pipeline architecture has four stages.

1. **Schema Analysis and Inference:** Raw ITSM exports are interpreted and normalized into a common schema using LLMs, accommodating variance in field names, formats, and completeness.
2. **Data Preprocessing Layer:** The text is preprocessed, and noise is reduced, including the removal of alert-like records that would otherwise dominate semantic clustering.
3. **Sales/Executive Topic Identification:** This stage is the core AI stage where ticket records are grouped using classical ML clustering algorithm - **Hierarchical Density-Based Spatial Clustering of Applications with Noise** (HDBSCAN) into first-level themes. These first-level themes are called **Sub-topics**. Following this, Sub-topics are aggregated into higher-level themes through hierarchical abstraction, using another clustering algorithm - **Hierarchical Agglomerative Clustering** (HAC). These second-level themes are called **Main-topics.** Both Sub-topics and Main-topics have been labelled using LLMs, post clustering. The LLM determines the label based on top 30 cluster members (closest to each cluster's centroid).
4. **UI Layer**: Provides a report with Main-topic to Sub-topic to ITSM ticket drilldowns. The report also has supporting information such as executive summary, ticket volume, MTTR patterns, etc.

## Methodology, Metrics, and Evaluation Criteria

The study follows DSR methodology (Hevner et al., 2004; Peffers et al., 2007). The dataset contains ticket title, description, comments, close notes, and timestamps across schema-variant exports. The pipeline has been applied to generate a complete report. In this paper, we focus more on metrics for RQ2 and RQ3, while metrics for RQ1 will be addressed in future work.

| RQ | Element | Metrics |
|---|---|---|
| RQ2 | Main-topic/Sub-topic | Main-topic Coherence, Main-topic Distinctiveness, Sub-topic Granular Fit, Sub-topic Coherence, |
| RQ3 | Final report as decision-support artifact | Interpretability, Actionability, Trust, Likelihood of use |

**Table 1. RQ-Element-Metric alignment**

For RQ2 and RQ3, a stakeholder evaluation was conducted involving six reports/artifacts reviewed by five users each across Sales Engineering and executive or customer-success roles. Each metric was rated on a scale of 1 to 5 using the rubrics in Table 2, while Figure 2 shows a sample of the survey instrument.

| TL Report | Interpretability | Actionability | Trust | Likelihood of Use | Main Topic Theme Coherence | Main Topic Theme Distinctiveness | Sub-topic Theme Granular Fit | Sub-topic Theme Coherence |
|---|---|---|---|---|---|---|---|---|
| Artifact 1 | 5 | 4 | 5 | 5 | 4 | 4 | 4 | 4 |
| Artifact 2 | 4 | 5 | 4 | 5 | 5 | 4 | 4 | 4 |
| Artifact 3 | 3 | 4 | 3 | 3 | 4 | 3 | 3 | 3 |
| Artifact 4 | 4 | 5 | 4 | 5 | 4 | 5 | 4 | 4 |
| Artifact 5 | 4 | 5 | 4 | 5 | 5 | 3 | 4 | 3 |
| Artifact 6 | 5 | 5 | 5 | 5 | 5 | 3 | 5 | 4 |

Figure 2 - Sample of Survey provided to the Stakeholders

| Metric and Definition | Rubric |
|---|---|

| **Interpretability:**<br>How easy it is to navigate, interpret, and extract value from the report, especially the flow from Main-topics → Sub-topics. | **Rubric**<br>1 - Very hard to follow; the hierarchy is confusing.<br>2 - Understandable only with considerable effort; important information is hard to find.<br>3 - Generally usable but requires better structure.<br>4 - Easy to interpret; I can find what I need without much effort.<br>5 - Very intuitive and efficient; I can move quickly from Main-topics to Sub-topics themes. |
|---|---|
| **Actionability:**<br>The extent to which the report helps the reviewer identify clear next steps for sales<br>conversations, executive prioritization, automation discussions, or customer success planning | **Rubric**<br>1 - The report is mostly descriptive; cannot identify clear next step. 2 - The report hints at some issues/opportunities, but next steps are vague or hard to derive.<br>3 - I can infer at least one reasonable next step, but I still need substantial extra thinking or analysis.<br>4 - The report supports several concrete next steps<br>5 - The report enables immediate, confident, and prioritized action. |
| **Trust:**<br>The degree to which the reviewer believes the report is credible, faithful to the ITSM data, and defensible. | **Rubric**<br>1 - I do not trust this report; several outputs seem unsupported. 2 - I have low trust; most labels or conclusions seem doubtful.<br>3 - The report is mostly plausible, but I would want to verify important claims before using it.<br>4 - I mostly trust the report; the outputs seem credible and usable. 5 - I strongly trust the report and use it in a real decision setting. |
| **Likelihood of Use:**<br>The likelihood that the reviewer actually uses the report for pre sales preparation, executive discussion, or customer success work. | **Rubric**<br>1 - I would not use it.<br>2 - I am unlikely to use it.<br>3 - I might use it occasionally in limited cases.<br>4 - I would likely use it regularly when relevant.<br>5 - I would actively rely on it as a standard input. |
| **Main-topic Theme Coherence:**<br>The degree to which the Sub topics under a Main-topic<br>belong together under one higher-level theme. | **Rubric**<br>1 - The grouped Sub-topics do not belong together.<br>2 - The grouping feels weak or forced; multiple Sub-topics seem misplaced.<br>3 - The grouping is mostly reasonable, but some boundary problems remain.<br>4 - The grouping is strong; most Sub-topics clearly fit together. 5 - The grouping is highly coherent; the Sub-topics clearly form one unified executive |

| **Main-topic Theme Distinctiveness:** How clearly a Main-topic is different from other Main-topics in the same report. | **Rubric**<br>1 - Heavily overlaps with another Main-topic; nearly duplicative. 2 - Substantial overlap; hard to explain how it differs from peer Main-topics.<br>3 - Somewhat distinct, but boundaries are still blurry. 4 - Clearly distinct from other Main-topics.<br>5 - Sharply differentiated and non-redundant. |
|---|---|
| **Sub-topic Theme Granular Fit:** Whether the Sub-topic is at the right specificity level for drill down, troubleshooting, and operational interpretation. | **Rubric**<br>1 - Far too broad or too fragmented to be useful.<br>2 - Somewhat mismatched in specificity.<br>3 - Acceptable, but not ideal.<br>4 - Good operational granularity.<br>5 - Ideal level of detail for drill-down. |
| **Sub-topic Theme Coherence:** The degree to which the tickets within the Sub-topic appear to reflect one underlying incident or request pattern. | **Rubric**<br>1 - The Sub-topic appears mixed or unrelated.<br>2 - Significant noise; multiple issue types seem mixed together. 3 - Mostly coherent, but some examples appear off-theme. 4 - Strongly coherent; the ticket pattern is clear.<br>5 - Very tight cluster; the ticket pattern is highly consistent |

**Table 2 - Definition and Rubrics (on a scale of 1 to 5) for Stakeholder Metrics. These were collected for evaluating Abstraction gap and Decision-support gap.**

Our approach operationalizes these metrics directly in embedding space. For each dataset, we generate embeddings for ticket titles (for Sub-topic metrics) and Sub-topics (for Main-topic metrics), to maintain a clear hierarchy. Within each Sub-topic cluster, we select the **top 30 tickets most semantically similar to the cluster label**, approximating what a human reviewer would examine. Coherence is then computed as the **average cosine similarity between items and their cluster centroid**, capturing how tightly grouped the cluster is. Distinctiveness is measured as the **distance between cluster centroids**, specifically how far each cluster is from its nearest neighbor, reflecting separation from other clusters. At the Main-topic level, the same methodology is applied using Sub-topic centroids, preserving the hierarchical structure. Cluster-level scores are then aggregated into global metrics for each artifact.

Because ML metrics are continuous while human ratings are on a **1 to 5 scale**, we apply **min-max normalization within each artifact** to align the scales and enable direct comparison. The methodology
intentionally favors simplicity and interpretability: we use fixed top-30 sampling to mirror human evaluation behavior, compute only global scores to match the level of human feedback, avoid weighting schemes, and exclude very small clusters where metrics are unstable. This approach ensures a clean, reproducible framework for assessing whether embedding-based measures meaningfully capture human perceptions of clustering quality.

For RQ3, the final report is treated as the decision-support artifact. Each metric is theoretically grounded: interpretability reflects the requirement that artifacts support easy navigation and repeated use (Brooke, 1996; Yigitbasioglu & Velcu, 2012); actionability is justified by task–technology fit, which holds that performance gains emerge only when outputs align with users' task requirements (Goodhue & Thompson, 1995); trust is treated as a distinct construct because it independently shapes adoption of AI-generated outputs (Glikson & Woolley, 2020); and likelihood of use serves as the behavioral intention outcome consistent with technology acceptance theory (Davis, 1989; Venkatesh et al., 2003).

## Preliminary Results and Discussion

**Evaluation of RQ1:** We assess the pipeline's capability to transform large-scale ITSM data into structured, decision-ready intelligence. The proposed process successfully standardizes heterogeneous ticket data and produces hierarchical abstractions (Main-topics and Sub-topics) across datasets containing up to 250,000 tickets in about 6 hours, under typical cloud-based processing conditions. Such large-scale abstraction is typically infeasible through manual analysis, particularly when requiring clustering, labeling, and hierarchical organization. While a controlled efficiency comparison is reserved for future work, this evidence demonstrates the pipeline's scalability and practical feasibility as a transformation mechanism.

| Artifact | Interp. | Action. | Trust | Lik. of Use | MT Coh. | MT Dist. | ST Gran. | ST Coh. |
|---|---|---|---|---|---|---|---|---|
| Artifact 1 | 4.0 | 4.2 | 4.4 | 4.8 | 4.0 | 4.0 | 4.2 | 4.0 |
| Artifact 2 | 3.4 | 4.0 | 4.0 | 3.6 | 3.8 | 4.0 | 3.4 | 4.2 |
| Artifact 3 | 4.4 | 4.4 | 4.6 | 4.4 | 4.4 | 3.8 | 3.8 | 4.2 |
| Artifact 4 | 4.2 | 4.0 | 4.0 | 4.2 | 4.4 | 3.8 | 4.0 | 4.4 |
| Artifact 5 | 4.2 | 4.2 | 4.2 | 4.0 | 4.2 | 3.8 | 4.0 | 3.4 |
| Artifact 6 | 5.0 | 4.6 | 4.8 | 4.6 | 4.4 | 4.6 | 4.2 | 4.0 |
| **Global Mean** | **4.20** | **4.23** | **4.33** | **4.27** | **4.20** | **4.00** | **3.93** | **4.03** |
| **Std** | **0.76** | **0.68** | **0.61** | **0.91** | **0.81** | **0.83** | **0.69** | **0.62** |

**Table 3. Human Stakeholder Ratings - RQ2 and RQ3 Metrics (Mean across 5 raters; scale 1–5)**
Interp. = Interpretability; Action. = Actionability; Lik. of Use = Likelihood of Use; MT Coh. = Main topic Theme Coherence; MT Dist. = Main-topic Theme Distinctiveness; ST Gran. = Sub-topic Granular Fit; ST Coh. = Sub-topic Theme Coherence. Std = sample standard deviation across all 30 ratings (5 raters × 6 artifacts).

**RQ2 - Abstraction Quality: Human Ratings:** Across all artifacts, Main-topic Coherence (4.20) and Sub-topic Coherence (4.03) score higher than Main-topic Distinctiveness (4.00) and Sub-topic Granular Fit (3.93). This coherence-over-distinctiveness pattern is observed across the majority of artifacts across our evaluation set and is an interesting finding in the RQ2 results. The pipeline appears to reliably produce internally consistent clusters across our evaluation set but is less effective at ensuring those clusters are sufficiently differentiated from one another. This finding directly motivates the abstraction-tuning experiments (details mentioned in Future Work).

**RQ2 - ML vs. Human Alignment:** Table 4 reports scaled ML scores alongside human averages for the four RQ2 metrics, with Pearson correlations computed across the six artifacts. The correlations reveal a metric-dependent alignment pattern with two distinct groups. Main-topic Distinctiveness (r = 0.78) and Sub-topic Coherence (r = 0.64) show meaningful alignment, indicating that for these metrics, embedding based scores may be directionally consistent with human judgment and could serve as scalable proxies, though larger samples are needed before treating them as reliable proxies. Main Topic Coherence (r = -0.12) and Sub-topic Granular Fit (r = -0.28) show no meaningful alignment. For Main-topic Coherence, human ratings cluster tightly between 3.8 and 4.4 while ML scores range widely from 1.88 to 4.60, suggesting that humans evaluate Main-topic coherence based on conceptual grouping logic and label quality rather than embedding geometry alone. For Granular Fit, the absence of alignment is theoretically expected: whether a Sub-topic is at the right specificity level for operational use is inherently a task-fit judgment (Goodhue & Thompson, 1995) that cosine similarity may not capture. These findings support the paper's hybrid evaluation logic that computational metrics are necessary but insufficient, and human judgment adds irreplaceable signal for certain abstraction dimensions in this context.

| Artifact | MT Coh. ML | MT Coh. H | MT Dist. ML | MT Dist. H | ST Gran. ML | ST Gran. H | ST Coh. ML | ST Coh. H |
|---|---|---|---|---|---|---|---|---|
| Artifact 1 | 1.88 | 4.0 | 2.09 | 4.0 | 2.25 | 4.2 | 4.19 | 4.0 |
| Artifact 2 | 4.60 | 3.8 | 2.80 | 4.0 | 2.89 | 3.4 | 4.04 | 4.2 |
| Artifact 3 | 2.50 | 4.4 | 2.68 | 3.8 | 2.88 | 3.8 | 3.31 | 4.2 |
| Artifact 4 | 3.14 | 4.4 | 2.06 | 3.8 | 1.46 | 4.0 | 3.80 | 4.4 |
| Artifact 5 | 3.49 | 4.2 | 2.59 | 3.8 | 2.68 | 4.0 | 2.56 | 3.4 |
| Artifact 6 | 4.60 | 4.4 | 3.54 | 4.6 | 3.00 | 4.2 | 4.25 | 4.0 |
| **r (ML, H)** | **-0.12** | | **0.78** | | **-0.28** | | **0.64** | |

**Table 4. ML Scaled Scores vs. Human Average Ratings - RQ2 Metrics (scale 1 to 5)**
ML = scaled ML score; H = human average; r = Pearson correlation across 6 artifacts. MT = Main topic; ST = Sub-topic; Coh. = Coherence; Dist. = Distinctiveness; Gran. = Granular Fit.

**RQ3 -Decision-Support Value:** Table 3 reports stakeholder ratings across all six artifacts and five raters. All four decision-support metrics exceed 4.0 globally, offering early-stage evidence, subject to the sample size limitations noted below, that the artifact is perceived as interpretable, actionable, credible, and adoption-worthy. Trust is the highest-rated metric (4.33) and the most consistent across raters (std = 0.61), suggesting that even when stakeholders found an artifact harder to navigate, they generally found its outputs credible. This aligns with prior work treating trust as a construct that can hold independently of usability (Glikson & Woolley, 2020). Likelihood of Use shows the widest spread (std = 0.91), indicating that adoption intent is more sensitive to artifact-specific quality variation than credibility or actionability. Artifact 6 is the strongest performer across all RQ3 metrics, with Interpretability reaching a ceiling score of 5.0 and high inter-rater agreement. Interestingly, Artifact 2 maintains a Trust score of 4.0 (despite having relatively lower Interpretability and Likelihood of Use), suggesting the finding that credibility and usability are separable dimensions.

**Systematic ML underestimation:** Across all six artifacts, ML-scaled scores for Main-topic Distinctiveness are uniformly lower than human ratings, with gaps ranging from 1.06 to 1.91 points. A similar pattern holds for Sub-topic Granular Fit (gaps of 0.51 to 2.54 points in five of six artifacts). This systematic underestimation suggests that the centroid-distance formulation is conservative. One possible explanation is that embedding space partially collapses conceptual distinctions that human evaluators detect through label semantics and thematic logic. Practically, this means ML distinctiveness scores may flag separation problems that humans do not perceive, making them more useful as a lower-bound signal than an absolute measure.

## Limitations and Future Work

This study has several limitations at its current stage.

1. The evaluation is performed on a small dataset with limited number of human reviewers/stakeholders, rather than full comparative testing.
2. The current evidence comes from one organizational setting, which limits claims about generalizability across firms or ITSM environments.
3. Reviewers may have had prior familiarity with some reports, which could introduce rating bias. Additionally, the six artifacts were selected as a convenience sample based on report availability and data integrity, not as a random or representative sample, limiting generalizability claims.

Before ICIS 2026, we will complete three extensions. First, we compute the metrics for RQ1, such as time to-insight and cost proxy. Second, increase the number of artifacts and the number of reviewers to achieve a more robust and reliable evaluation. Inter-rater reliability will be formally assessed as well because ICC analysis requires a larger artifact set (Koo & Li, 2016). Third, we will perform abstraction-tuning experiments across alternative clustering and hierarchy settings to identify the strongest trade-off among abstraction, coherence, and distinctiveness. This is based on our findings regarding **RQ2 - ML vs. Human Alignment.** Such tuning is theoretically important for balancing interpretability and information loss.

## Conclusion

This paper reframes ITSM data use as an IS problem of transformation, abstraction, and decision support. The implemented AI pipeline converts heterogeneous ticket exports into a multilevel, decision-ready report through schema normalization, HDBSCAN-based sub-topic clustering, and HAC-based main-topic abstraction. Preliminary evaluation across six artifacts and five stakeholders shows that all four decision support metrics exceed 4.0, with Trust emerging as the most consistent signal (mean = 4.33, std = 0.61), providing early evidence that the artifact is perceived as credible and adoption-worthy. In doing so, this paper transforms raw ITSM tickets into high-fidelity insights for sales and executive decision making.